\documentclass[runningheads]{llncs}
\usepackage[T1]{fontenc}
\usepackage{graphicx,verbatim}
\usepackage{hyperref}
\usepackage{color}
\usepackage{arydshln}
\usepackage{booktabs}
\usepackage[final]{microtype}
\usepackage{xurl} % better URL line breaks in references
\begin{document}
\title{Multi-Teacher Contrastive Distillation for Edge-Efficient Pathology Foundation Models}
\titlerunning{MuCoDi for Edge-Efficient Pathology Foundation Models}
%\begin{comment} %% Removed for anonymized MICCAI submission
\author{Tim Lenz~\inst{1}{$^\dagger$} \and 
Maurice Heide~\inst{1}{$^\dagger$} \and 
Marco Gustav~\inst{1} \and
Nic G. Reitsam~\inst{1,2,3} \and \\
Jakob Nikolas Kather~\inst{1,4,5,6}{*}
}
\authorrunning{T. Lenz and M. Heide et al.}
\institute{EKFZ for Digital Health, TU Dresden, Dresden, Germany\\
\and
Pathology, University of Augsburg, Augsburg, Germany
\and
Bavarian Cancer Research Center (BZKF), Augsburg, Germany
\and
Department of Medicine I, TU Dresden, Dresden, Germany
\and 
Medical Oncology, NCT Heidelberg, Heidelberg, Germany
\and 
Pathology \& Data Analytics, University of Leeds, Leeds, United Kingdom \\
\email{kather.jn@tu-dresden.de}
}
%\end{comment}
 
\maketitle       % typeset the header of the contribution
\begin{abstract}
Computational pathology foundation models (PFMs) have advanced whole-slide image analysis. However, their size and inference cost hinder local deployment in pathology departments. We propose MuCoDi, a pretraining framework that distills frozen tile embeddings from multiple PFMs into compact edge-oriented encoders. Instead of regressing individual teacher features, MuCoDi trains lightweight MobileOne and RepViT students with a contrastive distillation objective adapted from MoCo v3, where cached Virchow2, UNI2, and H-Optimus-1 embeddings replace momentum-encoder keys. We pretrain students on 14.3M TCGA tiles from only 11.8K WSIs and evaluate frozen encoders on 23 clinically curated downstream classification tasks. RepViT-based MuCoEdge students retain near-teacher performance while reducing model size by orders of magnitude: MuCoEdge-R2.3 and MuCoEdge-R1.5 reach 71.0\% external AUROC, within 0.8 percentage points of the best teacher (Virchow2, 71.8\%), while MuCoEdge-R2.3 obtains the best external F1 and the second-best AUPRC (51.8\% and 53.3\%). MuCoEdge-R1.0 reaches 70.9\% AUROC with only 6.4M parameters and 1.12 GFLOPs. On a Raspberry Pi 5, sub-million-parameter MobileOne students achieve up to 605$\times$ single-tile speedup over Virchow2 while retaining 66.5--66.9\% external AUROC, demonstrating that PFM-quality pathology representations can be moved toward practical edge deployment. Code is available at \url{https://anonymous.4open.science/r/mucodi-6243}.
\end{abstract}
\let\thefootnote\relax\footnotetext{$^\dagger$~Equal contribution, {*} Corresponding author}
\begin{figure}
  \centering
    \includegraphics[width=0.9\linewidth]{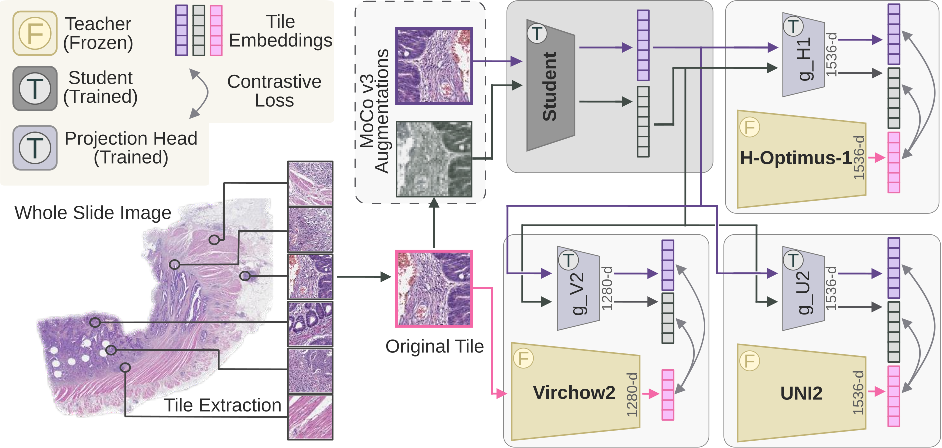}
  \caption{Overview of MuCoDi pretraining. Frozen PFM teachers provide cached tile embeddings, while a lightweight student encoder and teacher-specific projection heads are trained with a multi-teacher contrastive distillation loss.}
  \label{fig:overview}
\end{figure}
\section{Introduction}
The central promise of computational pathology is to bring AI assistance to pathologists, rather than asking pathologists to adapt to AI infrastructure. This is increasingly urgent: global cancer incidence was close to 20 million new cases in 2022 and is projected to reach 35 million by 2050~\cite{bray2024global}. At the same time, histopathology workforces are under pressure from rising workload, increasing diagnostic complexity and aging workforces~\cite{walsh2024workforce}. AI systems could support screening, triage, and prescreening for therapy-relevant biomarkers, but only if they integrate into routine workflows without additional operational burden.

Current pathology foundation models (PFMs) have changed the performance landscape but not yet the deployment landscape. Their large vision-transformer backbones and gigapixel whole-slide image (WSI) pipelines often require dedicated GPUs, local clusters, or cloud processing. This creates a workflow mismatch: instead of augmenting existing pathology practice, many methods require additional digitization, transfer, preprocessing, and compute steps. For clinical adoption, models should run where tissue is examined and where data are produced: on existing pathology workstations, ordinary clinic computers, laptops, or eventually device-integrated microscopes. Such local inference also reduces data movement, allowing sensitive images to remain inside the pathology department.

Recent work highlights the feasibility and importance of this direction. Choudhury et al. developed a low-cost, open-source microscopy platform for slide capture and computational analysis, including Raspberry-Pi-based local compute for automated histopathology evaluation~\cite{choudhury2024lowcost}. LitePath similarly argues that PFM accessibility depends not only on accuracy but also on deployability, reporting large reductions in parameters, FLOPs, energy, and runtime through distilled LiteFM models and adaptive patch selection~\cite{cai2026litepath}. These studies motivate a practical target for pathology AI: compact encoders that can run inside existing departmental infrastructure, including low-cost workstation and microscope-connected edge setups, rather than requiring external high-performance compute.

We propose Multi-teacher Contrastive Distillation (MuCoDi), a framework for learning edge-efficient pathology encoders (Fig.~\ref{fig:overview}). MuCoDi distills complementary representations from multiple state-of-the-art PFMs into a substantially smaller student backbone designed for local inference. Instead of matching only individual teacher features, MuCoDi preserves teacher-induced relational structure through contrastive distillation, enabling a compact model to inherit complementary morphology-aware embeddings while reducing latency and hardware demands.

Our contributions are threefold. First, we introduce an efficient contrastive multi-teacher knowledge distillation pretraining framework for edge-oriented pathology representation learning using only 11.8K unlabeled TCGA WSIs for student pretraining. Second, we demonstrate extreme latency improvements while maintaining strong downstream classification performance, including sub-million-parameter students for severe edge constraints. Third, we benchmark the latency-accuracy trade-off of compact pathology encoders on low-cost edge hardware, targeting future integration into routine workstation and microscope-based pathology workflows.

\section{Related work}
\begin{table}[t]
\centering
\small
\caption{Overview of feature extractors in this study. Abbreviations: \#Ps denotes parameters in millions, GFLOPs denotes giga floating-point operations for one ($224\times224$)~px tile, and \#WSI denotes pretraining whole-slide images in thousands. The asterisk marks the custom ViT-G/14 configuration used by UNI2.}
\label{tab:feature_extractor_overview}
\setlength{\tabcolsep}{3pt}
\begin{tabular}{l l r r r}
\hline
Model & Backbone & \#Ps[M] & GFLOPs & \#WSI[K] \\
\hline
H-Optimus-1~\cite{scalbert2026hoptimus1} & ViT-G/14 & 1135.0 & 295.97 & $>$1000 \\
UNI2~\cite{mahmoodlab2025uni2} & ViT-G/14$^{*}$ & 681.0 & 180.39 & $>$350 \\
Virchow2~\cite{zimmermann2024virchow2} & ViT-H/14 & 632.0 & 164.59 & 3100 \\
LiteFM-L~\cite{cai2026litepath} & ViT-B & 86.6 & 17.58 & 72.3 \\
LiteFM~\cite{cai2026litepath} & ViT-S & 22.1 & 4.61 & 72.3 \\
LiteFM-S~\cite{cai2026litepath} & ViT-T & 5.7 & 1.26 & 72.3 \\
\hline
MuCoEdge-R2.3 & RepViT-M2.3 & 22.4 & 4.57 & 11.8 \\
MuCoEdge-R1.5 & RepViT-M1.5 & 13.6 & 2.31 & 11.8 \\
MuCoEdge-Ms3 & MobileOne-S3 & 8.0 & 1.89 & 11.8 \\
MuCoEdge-R1.1 & RepViT-M1.1 & 7.8 & 1.36 & 11.8 \\
MuCoEdge-R1.0 & RepViT-M1.0 & 6.4 & 1.12 & 11.8 \\
MuCoEdge-Ms2 & MobileOne-S2 & 5.8 & 1.30 & 11.8 \\
MuCoEdge-R0.9 & RepViT-M0.9 & 4.7 & 0.831 & 11.8 \\
MuCoEdge-Ms1 & MobileOne-S1 & 3.5 & 0.824 & 11.8 \\
MuCoEdge-R0.6 & RepViT-M0.6 & 2.2 & 0.394 & 11.8 \\
MuCoEdge-Ms0 & MobileOne-S0 & 1.1 & 0.274 & 11.8 \\
MuCoEdge-M$\mu$2 & MobileOne-$\mu$2 & 0.7 & 0.214 & 11.8 \\
MuCoEdge-M$\mu$1 & MobileOne-$\mu$1 & 0.5 & 0.139 & 11.8 \\
MuCoEdge-M$\mu$0 & MobileOne-$\mu$0 & 0.2 & 0.069 & 11.8 \\
\hline
\end{tabular}
\end{table}
\paragraph{Pathology foundation models.}
Large self-supervised pathology foundation models (PFMs) such as UNI~\cite{chen2024uni}, Virchow~\cite{vorontsov2024virchow}, Prov-GigaPath~\cite{xu2024gigapath}, H-optimus-0/1~\cite{saillard2024hoptimus0,scalbert2026hoptimus1}, and Virchow2~\cite{zimmermann2024virchow2} have substantially improved reusable histopathology representations. However, this progress largely follows a scaling trend: recent models range from hundreds of millions to over one billion parameters and process whole-slide images through large numbers of tiles, making training and deployment costly~\cite{neidlinger2026benchmarking}.

\paragraph{Knowledge distillation in pathology.}
Knowledge distillation (KD) compresses teacher models into a student~\cite{hinton2015distilling}. Contrastive representation distillation instead transfers relational structure in feature space~\cite{tian2020crd}. In computational pathology, multi-model knowledge has been used to build stronger or more task-adapted representations: GPFM combines expert and self-KD~\cite{ma2026gpfm}, COBRA treats embeddings from multiple PFMs as feature-space augmentations for slide-level contrastive learning~\cite{lenz2025cobra}, and HistoMILKD distills multiple PFMs into a MIL-based WSI classifier~\cite{mallya2026histomilkd}. These methods demonstrate the value of multi-PFM supervision, but they primarily target generalization or slide-level prediction rather than compact feature extractors for efficient edge inference.

\paragraph{Efficient pathology foundation models.}
Recent work has begun to address PFM efficiency. H0-mini is an 86M-parameter ViT-B/14 distilled from a large pathology FM for robust and efficient inference~\cite{filiot2025h0mini}. Virchow2G Mini is a 22M-parameter distillation of the 1.9B-parameter Virchow2G~\cite{zimmermann2024virchow2}. LitePath introduces LiteFM, a multi-teacher distilled ViT family with adaptive patch selection for resource-efficient WSI analysis~\cite{cai2026litepath}. Nevertheless, these models remain ViT-based and are not primarily designed or benchmarked for direct deployment on low-cost edge devices integrated into routine pathology hardware.

\paragraph{Efficient architectures.}
In general computer vision, mobile-first backbones explicitly optimize latency and hardware efficiency. MobileNetV3 uses hardware-aware search~\cite{howard2019mobilenetv3}, MobileOne targets sub-millisecond mobile inference through reparameterized convolutional design~\cite{vasu2023mobileone}, RepViT revisits mobile CNNs through ViT-inspired design~\cite{wang2024repvit}, and EfficientFormer shows that transformer-like models can approach MobileNet-speed inference~\cite{li2022efficientformer}. MuCoDi brings this design philosophy to computational pathology by distilling multiple large PFMs into edge-oriented students for efficient integration into microscope or workstation workflows.

\section{Methods}

\subsection{MuCoDi pretraining}
MuCoDi adapts the two-view contrastive learning recipe of MoCo v3~\cite{chen2021mocov3} to multi-teacher distillation. MoCo v3 contrasts queries against keys produced by a momentum encoder. MuCoDi keeps the contrastive objective but replaces this key encoder with cached embeddings from frozen pathology foundation models, which act as fixed teacher keys. For each training tile $x$, we precompute teacher features $k_t(x)$ for $T=3$ teachers: Virchow2, UNI2, and H-Optimus-1. The teachers remain frozen. Only the lightweight student backbone $f_\theta$ and teacher-specific linear projection heads $g_t$ are optimized. The heads map the shared student feature to 1280 dimensions for Virchow2 and 1536 dimensions for UNI2 and H-Optimus-1.
For each tile, we sample two MoCo-v3-style stochastic views $x_1,x_2$ resized to ($224\times224$)~px. Let $q_{t,v}=\mathrm{norm}(g_t(f_\theta(x_v)))$ be the student query for teacher $t$ and view $v\in\{1,2\}$. Let $K_t=\{k_t^{(i)}\}_{i=1}^{B_g}$ denote the L2-normalized teacher keys gathered across all GPUs for the current global batch of size $B_g$. For a query from tile $j$, the positive key $k_t^{+(j)}$ is the cached teacher embedding of the same underlying tile. For one teacher and view, the InfoNCE term~\cite{oord2018cpc} is
\begin{equation}
\ell_{t,v}^{(j)} = -\log
\frac{\exp\left(q_{t,v}^{(j)\top} k_t^{+(j)} / \tau\right)}
{\sum_{i=1}^{B_g} \exp\left(q_{t,v}^{(j)\top} k_t^{(i)} / \tau\right)} ,
\end{equation}
where $\tau$ is the temperature. Following MoCo v3, the optimized per-sample loss is scaled as $\mathcal{L}_{t,v}^{(j)}=2\tau\ell_{t,v}^{(j)}$, which makes the gradient magnitude independent of temperature. The final MuCoDi objective is the unweighted sum over teachers and views,
\begin{equation}
\mathcal{L}_{\mathrm{MuCoDi}} = \sum_{t=1}^{T}\sum_{v=1}^{2}\frac{1}{B_g}\sum_{j=1}^{B_g}\mathcal{L}_{t,v}^{(j)} .
\end{equation}
Teacher embeddings are stored with the image tiles and loaded without gradients, and negatives are provided by cross-GPU gathering of the current batch.

\paragraph{Training details.}
In the student sweep, all students are pretrained for 10 epochs with AdamW using PyTorch default betas, base learning rate $2\times10^{-2}$, weight decay $10^{-6}$, global batch size 2048, and temperature $\tau=0.2$. The learning rate is linearly warmed up for the first 5\% of optimizer steps and then cosine-decayed to zero. Gradients are clipped to a maximum norm of 1.0. Training uses bfloat16 automatic mixed precision, SyncBatchNorm, and distributed data parallelism on four GPUs. The augmentation pipeline follows the MoCo v3 two-crop scheme with random resized crops, color jitter, grayscale conversion, horizontal flips, asymmetric Gaussian blur, and solarization.
\subsection{Data and task selection}
MuCoDi is pretrained on TCGA tissue tiles with cached teacher features~\cite{weinstein2013tcga}. The pretraining set comprises 11,803 FFPE diagnostic WSIs and 14,280,892 tissue tiles. As shown in Tab.~\ref{tab:feature_extractor_overview}, this student distillation corpus uses approximately 6$\times$ fewer WSIs than LiteFM and over 260$\times$ fewer than Virchow2. Pretraining and evaluation use the same 2.0~$\mu$m/px (5$\times$) physical resolution, which reduces tile count and compute~\cite{neidlinger2025eagle}. Downstream evaluation follows the weakly supervised tile-feature benchmarking paradigm used in recent slide-representation and pathology foundation-model studies~\cite{lenz2025cobra,neidlinger2026benchmarking}. 
Task heads are trained on TCGA and deployed unchanged on matched CPTAC cohorts~\cite{ellis2013cptac}. Evaluation uses the matched TCGA set (3,933 patients, 4,633 slides) and matched CPTAC set (1,307 patients, 3,736 deployable slides), covering 23 clinically curated binary endpoints with at least 20 minority-class patients. All three teacher PFMs achieve at least 0.65 AUROC on internal TCGA evaluation for each endpoint, ensuring that each endpoint has a morphology-predictable signal before external deployment. The selected tasks span BRCA (\textit{TP53}, \textit{ESR1}, and \textit{PGR}), CRC (\textit{TP53} and MSI), GBM (\textit{NF1}), KIRC (\textit{SETD2}, grade, stage, and T stage), LUAD (\textit{EGFR}, \textit{STK11}, \textit{TP53}, and T stage), NSCLC subtype, and UCEC (\textit{BRD4}, \textit{KMT2A}, \textit{MGA}, \textit{NF1}, \textit{PTEN}, \textit{ROS1}, \textit{TP53}, and grade).

\subsection{Evaluation protocol}
Downstream performance is measured by freezing the student as a tile encoder, extracting one feature vector per tile, and training slide-level multiple-instance learning (MIL) heads with the STAMP workflow~\cite{elnahhas2025stamp} on TCGA using patient-level stratified 5-fold cross-validation. External generalization is evaluated by applying each TCGA-trained fold checkpoint unchanged to the matched CPTAC cohort.

\paragraph{Slide-level heads and metrics.}
Main results use the STAMP transformer MIL head. Performance is reported as AUROC, F1 score, and AUPRC with 95\% confidence-interval half-widths.

\paragraph{Edge inference benchmark.}
Deployment efficiency is measured on a Raspberry Pi 5 Model B with a quad-core Cortex-A76 CPU, 16~GB RAM, and no inference accelerator. Latency is measured in single-threaded float32 CPU mode without quantization and with batch size 1. MobileOne and RepViT models are first converted to their fused inference-time forms. For each model, we report parameters, GFLOPs for one ($224\times224$)~px RGB tile, and median single-tile latency from 100 timed forward passes after 20 warmup passes.

\section{Results}

\begin{table}
\centering
\scriptsize
\setlength{\tabcolsep}{4pt}
\resizebox{\textwidth}{!}{%
\providecommand{\hdashline}{\midrule}
\begin{tabular}{l r l c c c | c c c}
\toprule
Model & Params & Type & \multicolumn{3}{c}{TCGA (internal)} & \multicolumn{3}{c}{CPTAC (external)} \\
\cmidrule(lr){4-6} \cmidrule(lr){7-9}
 & & & AUC & F1 & AUPRC & AUC & F1 & AUPRC \\
\midrule
H-Optimus-1 & 1135.0M & Theirs & 74.1\textsubscript{7.7} & 53.4\textsubscript{8.1} & 55.4\textsubscript{8.7} & 68.9\textsubscript{3.6} & 48.1\textsubscript{5.7} & 51.1\textsubscript{3.0} \\
UNI2 & 681.0M & Theirs & 76.4\textsubscript{7.1} & 55.3\textsubscript{7.9} & \textbf{58.8\textsubscript{8.9}} & \underline{71.7\textsubscript{3.3}} & 50.6\textsubscript{6.1} & \textbf{53.4\textsubscript{3.2}} \\
Virchow2 & 632.0M & Theirs & 76.8\textsubscript{7.8} & \textbf{56.7\textsubscript{7.8}} & 58.7\textsubscript{8.5} & \textbf{71.8\textsubscript{3.6}} & 50.7\textsubscript{6.8} & 52.9\textsubscript{3.8} \\
\hline
LiteFM-L & 86.6M & Theirs & 75.0\textsubscript{6.7} & 53.9\textsubscript{7.6} & 56.8\textsubscript{8.1} & 69.9\textsubscript{3.1} & 48.8\textsubscript{6.3} & 51.7\textsubscript{3.1} \\
LiteFM & 22.1M & Theirs & 74.0\textsubscript{8.6} & 53.1\textsubscript{7.7} & 55.8\textsubscript{8.7} & 69.0\textsubscript{3.4} & 47.4\textsubscript{5.8} & 50.9\textsubscript{3.4} \\
LiteFM-S & 5.7M & Theirs & 73.4\textsubscript{8.5} & 52.8\textsubscript{8.2} & 54.8\textsubscript{8.2} & 68.3\textsubscript{3.9} & 46.7\textsubscript{6.5} & 50.2\textsubscript{3.7} \\
\hline
MuCoEdge-R2.3 & 22.4M & Ours & \textbf{77.1\textsubscript{7.3}} & 55.6\textsubscript{6.3} & 58.5\textsubscript{7.8} & 71.0\textsubscript{3.2} & \textbf{51.8\textsubscript{4.8}} & \underline{53.3\textsubscript{3.4}} \\
MuCoEdge-R1.5 & 13.6M & Ours & 76.7\textsubscript{7.3} & 55.4\textsubscript{7.9} & 57.8\textsubscript{8.7} & 71.0\textsubscript{2.9} & 51.1\textsubscript{5.7} & 52.8\textsubscript{3.1} \\
MuCoEdge-R1.1 & 7.8M & Ours & 76.3\textsubscript{7.6} & 55.2\textsubscript{7.7} & 58.3\textsubscript{9.1} & 70.9\textsubscript{3.3} & 50.5\textsubscript{5.4} & 52.2\textsubscript{3.4} \\
MuCoEdge-R1.0 & 6.4M & Ours & \underline{76.9\textsubscript{7.7}} & 55.6\textsubscript{8.0} & \underline{58.7\textsubscript{9.4}} & 70.9\textsubscript{2.9} & 50.2\textsubscript{6.3} & 52.0\textsubscript{3.4} \\
MuCoEdge-R0.9 & 4.7M & Ours & 76.6\textsubscript{8.0} & \underline{56.2\textsubscript{7.4}} & 58.6\textsubscript{9.6} & 70.5\textsubscript{3.3} & \underline{51.3\textsubscript{5.0}} & 52.4\textsubscript{3.6} \\
MuCoEdge-R0.6 & 2.2M & Ours & 75.2\textsubscript{8.0} & 54.0\textsubscript{7.2} & 57.3\textsubscript{9.3} & 69.4\textsubscript{3.7} & 48.8\textsubscript{5.8} & 51.1\textsubscript{4.0} \\
\hdashline
MuCoEdge-Ms3 & 8.0M & Ours & 74.6\textsubscript{7.6} & 54.0\textsubscript{7.9} & 56.6\textsubscript{8.8} & 68.7\textsubscript{3.6} & 46.4\textsubscript{5.1} & 50.9\textsubscript{3.8} \\
MuCoEdge-Ms2 & 5.8M & Ours & 74.9\textsubscript{7.7} & 53.6\textsubscript{7.6} & 56.7\textsubscript{9.2} & 69.3\textsubscript{3.6} & 46.8\textsubscript{6.2} & 51.1\textsubscript{3.6} \\
MuCoEdge-Ms1 & 3.5M & Ours & 74.6\textsubscript{7.1} & 53.7\textsubscript{8.1} & 55.8\textsubscript{7.7} & 68.5\textsubscript{3.9} & 46.0\textsubscript{5.7} & 50.7\textsubscript{3.7} \\
MuCoEdge-Ms0 & 1.1M & Ours & 73.2\textsubscript{9.2} & 52.6\textsubscript{8.2} & 54.9\textsubscript{8.3} & 67.4\textsubscript{4.3} & 45.2\textsubscript{7.6} & 49.7\textsubscript{3.8} \\
MuCoEdge-M$\mu$2 & 0.7M & Ours & 73.1\textsubscript{8.7} & 52.5\textsubscript{8.5} & 54.8\textsubscript{8.2} & 66.8\textsubscript{4.2} & 44.3\textsubscript{6.2} & 48.9\textsubscript{3.8} \\
MuCoEdge-M$\mu$1 & 0.5M & Ours & 72.9\textsubscript{8.4} & 52.2\textsubscript{7.1} & 54.4\textsubscript{8.0} & 66.9\textsubscript{3.9} & 45.7\textsubscript{5.6} & 48.6\textsubscript{3.5} \\
MuCoEdge-M$\mu$0 & 0.2M & Ours & 72.1\textsubscript{8.1} & 51.6\textsubscript{8.0} & 53.7\textsubscript{7.8} & 66.5\textsubscript{4.5} & 44.5\textsubscript{6.5} & 48.4\textsubscript{3.4} \\
\bottomrule
\end{tabular}
}
\caption{Aggregated results for TCGA (internal) and CPTAC (external) evaluation using the STAMP transformer MIL head. Values are means across the 23 selected comparison tasks and are reported in \%. Subscripts indicate the mean 95\% t-interval CI half-width across the five cross-validation folds. \textbf{Bold} and \underline{underlining} denote the best and second-best values in each column, respectively.}
\label{tab:agg_vit}
\end{table}

\begin{table}
\centering
\scriptsize
\setlength{\tabcolsep}{3pt}
\resizebox{\textwidth}{!}{%
\providecommand{\hdashline}{\midrule}
\begin{tabular}{l | c c | c c | c c | c c | c c | c c | c c}
\toprule
Model & \multicolumn{2}{|c|}{Breast} & \multicolumn{2}{c|}{Colon} & \multicolumn{2}{c|}{Brain} & \multicolumn{2}{c|}{Kidney} & \multicolumn{2}{c|}{Lung} & \multicolumn{2}{c|}{Uterus} & \multicolumn{2}{c}{\textbf{Avg}} \\
\cmidrule(lr){2-13} \cmidrule(lr){14-15}
 & AUC & F1 & AUC & F1 & AUC & F1 & AUC & F1 & AUC & F1 & AUC & F1 & AUC & F1 \\
\midrule
H-Optimus-1 & 78.7 & \underline{72.6} & 73.0 & 50.6 & 57.5 & 26.1 & 63.5 & 44.4 & 77.6 & 57.5 & 63.0 & 37.0 & 68.9 & 48.1 \\
UNI2 & \underline{81.2} & 69.5 & \underline{77.0} & 59.9 & 63.1 & 32.3 & \underline{66.4} & \textbf{47.9} & 77.4 & 57.8 & 66.9 & 40.2 & \underline{71.7} & 50.6 \\
Virchow2 & \textbf{83.3} & \textbf{76.4} & \textbf{78.0} & 60.6 & 60.2 & 28.4 & \textbf{66.9} & 44.9 & 75.9 & 56.8 & \underline{67.3} & 40.4 & \textbf{71.8} & 50.7 \\
\hline
LiteFM-L & 80.1 & 72.4 & 73.6 & 58.4 & 62.1 & \textbf{34.7} & 65.1 & 42.3 & 76.6 & 57.2 & 64.3 & 37.5 & 69.9 & 48.8 \\
LiteFM & 79.8 & 71.3 & 71.9 & 52.9 & 62.3 & 32.7 & 64.5 & 41.9 & 74.6 & 55.6 & 63.9 & 36.6 & 69.0 & 47.4 \\
LiteFM-S & 77.5 & 70.3 & 68.8 & 45.3 & 61.1 & 28.0 & 64.3 & 42.0 & 73.9 & 54.6 & 64.2 & 37.8 & 68.3 & 46.7 \\
\hline
MuCoEdge-R2.3 & 78.1 & 72.3 & 76.3 & \textbf{67.8} & 60.9 & 30.5 & 66.4 & 45.9 & \underline{78.0} & 58.5 & 66.1 & \underline{41.5} & 71.0 & \textbf{51.8} \\
MuCoEdge-R1.5 & 77.7 & 71.5 & 73.5 & 60.1 & 62.0 & 30.6 & 65.9 & 45.3 & \textbf{78.4} & 58.3 & 67.0 & \textbf{42.3} & 71.0 & 51.1 \\
MuCoEdge-R1.1 & 76.0 & 69.8 & 76.3 & 62.5 & 62.8 & 28.3 & 65.8 & 46.6 & 77.2 & \underline{58.7} & \textbf{67.3} & 40.0 & 70.9 & 50.5 \\
MuCoEdge-R1.0 & 77.8 & 70.6 & 74.8 & 55.8 & 60.9 & \underline{33.8} & 66.2 & 45.4 & 77.0 & 57.4 & 67.1 & 41.1 & 70.9 & 50.2 \\
MuCoEdge-R0.9 & 77.0 & 71.7 & 74.4 & \underline{63.3} & 60.3 & 32.8 & 65.8 & \underline{47.3} & 77.6 & \textbf{59.0} & 66.4 & 40.1 & 70.5 & \underline{51.3} \\
MuCoEdge-R0.6 & 73.8 & 67.0 & 73.1 & 54.6 & 60.6 & 31.2 & 65.0 & 45.4 & 76.5 & 57.4 & 65.6 & 39.1 & 69.4 & 48.8 \\
\hdashline
MuCoEdge-Ms3 & 77.9 & 68.6 & 70.3 & 40.7 & 63.2 & 31.1 & 64.1 & 43.8 & 74.6 & 54.8 & 64.2 & 37.4 & 68.7 & 46.4 \\
MuCoEdge-Ms2 & 77.7 & 64.9 & 73.4 & 52.3 & 62.8 & 28.2 & 65.1 & 43.0 & 74.8 & 55.1 & 64.6 & 37.6 & 69.3 & 46.8 \\
MuCoEdge-Ms1 & 78.1 & 66.6 & 70.9 & 48.8 & 62.2 & 31.0 & 65.2 & 42.4 & 74.2 & 54.6 & 63.1 & 36.0 & 68.5 & 46.0 \\
MuCoEdge-Ms0 & 75.7 & 58.8 & 71.0 & 54.5 & 62.2 & 30.3 & 64.1 & 41.9 & 72.5 & 52.6 & 62.6 & 36.5 & 67.4 & 45.2 \\
MuCoEdge-M$\mu$2 & 75.1 & 61.0 & 67.1 & 42.1 & \textbf{63.9} & 29.3 & 64.2 & 43.0 & 70.7 & 51.9 & 62.8 & 36.3 & 66.8 & 44.3 \\
MuCoEdge-M$\mu$1 & 74.9 & 61.4 & 67.8 & 53.4 & \underline{63.8} & 31.6 & 64.0 & 44.8 & 71.7 & 53.8 & 62.3 & 35.2 & 66.9 & 45.7 \\
MuCoEdge-M$\mu$0 & 73.0 & 58.4 & 68.4 & 51.1 & 61.4 & 28.3 & 64.4 & 43.1 & 71.0 & 51.9 & 62.4 & 35.8 & 66.5 & 44.5 \\
\bottomrule
\end{tabular}
}
\caption{Per-organ AUC and F1, CPTAC (external validation), STAMP transformer MIL head. Each cell is the mean over the organ's tasks (Lung pools LUAD and NSCLC); values in \%. \textbf{Bold} and \underline{underline} mark the best and second-best value per column.}
\label{tab:perorgan_vit_cptac}
\end{table}

\paragraph{External validation.}
We compare MuCoEdge students against their teacher PFMs and LiteFM competitors on the 23 selected downstream classification tasks (Tab.~\ref{tab:agg_vit}). Teacher PFMs retain the highest external AUROC, with Virchow2 reaching 71.8\% and UNI2 71.7\%. RepViT-based students close most of this gap at much smaller scale: MuCoEdge-R2.3 and MuCoEdge-R1.5 reach 71.0\% AUROC, while MuCoEdge-R1.1 and MuCoEdge-R1.0 reach 70.9\%. The strongest edge trade-off is achieved by smaller RepViT students. MuCoEdge-R0.9 reaches 70.5\% AUROC, 51.3\% F1, and 52.4\% AUPRC with 4.7M parameters, outperforming LiteFM-L on all three external metrics despite using 18$\times$ fewer parameters and 21$\times$ fewer GFLOPs. MuCoEdge-R0.6 remains competitive with LiteFM while using only 2.2M parameters. These results are obtained from only 11.8K student-pretraining WSIs, approximately 6$\times$ fewer than LiteFM and over 260$\times$ fewer than Virchow2 (Tab.~\ref{tab:feature_extractor_overview}). LiteFM also includes both TCGA and CPTAC in its pretraining corpus. At the extreme edge end, sub-million MobileOne variants retain 66.5--66.9\% external AUROC with only 0.2--0.7M parameters. Internal TCGA evaluation shows the same trend: MuCoEdge-R2.3 obtains the highest mean AUROC (77.1\%), and several students remain within the teacher range. Per-organ results (Tab.~\ref{tab:perorgan_vit_cptac}) show that performance is not driven by a single tissue type. MuCoEdge variants obtain the best average F1 and are competitive across colon, kidney, lung, and uterine tasks.

\paragraph{Efficiency and edge deployment.}
\begin{figure}[h!]
  \centering
  \setlength{\tabcolsep}{0pt}
  \begin{tabular}{@{}c@{\hspace{0.35cm}}c@{}}
    \includegraphics[height=4.45cm,keepaspectratio]{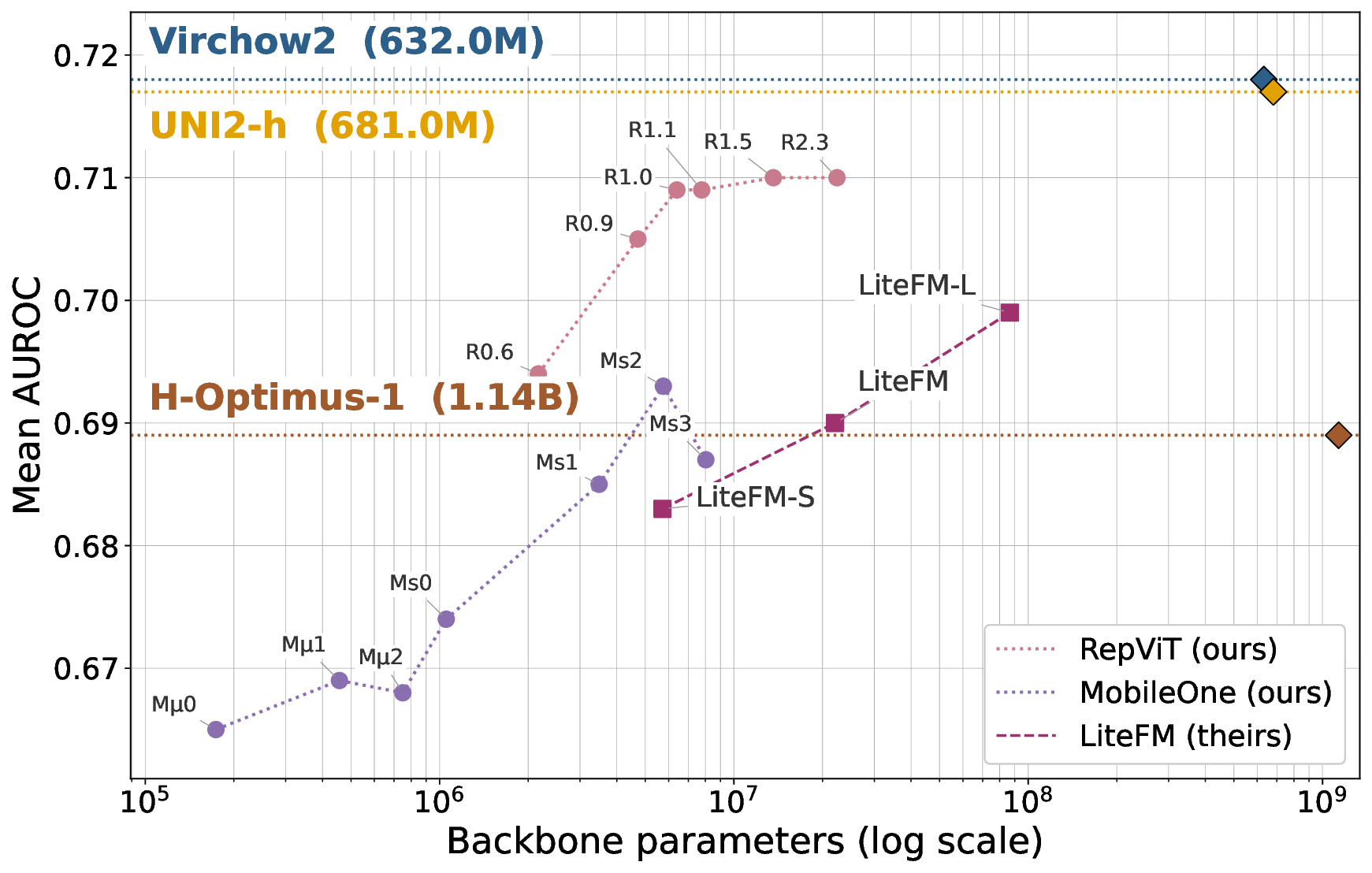} &
    \includegraphics[height=4.45cm,keepaspectratio]{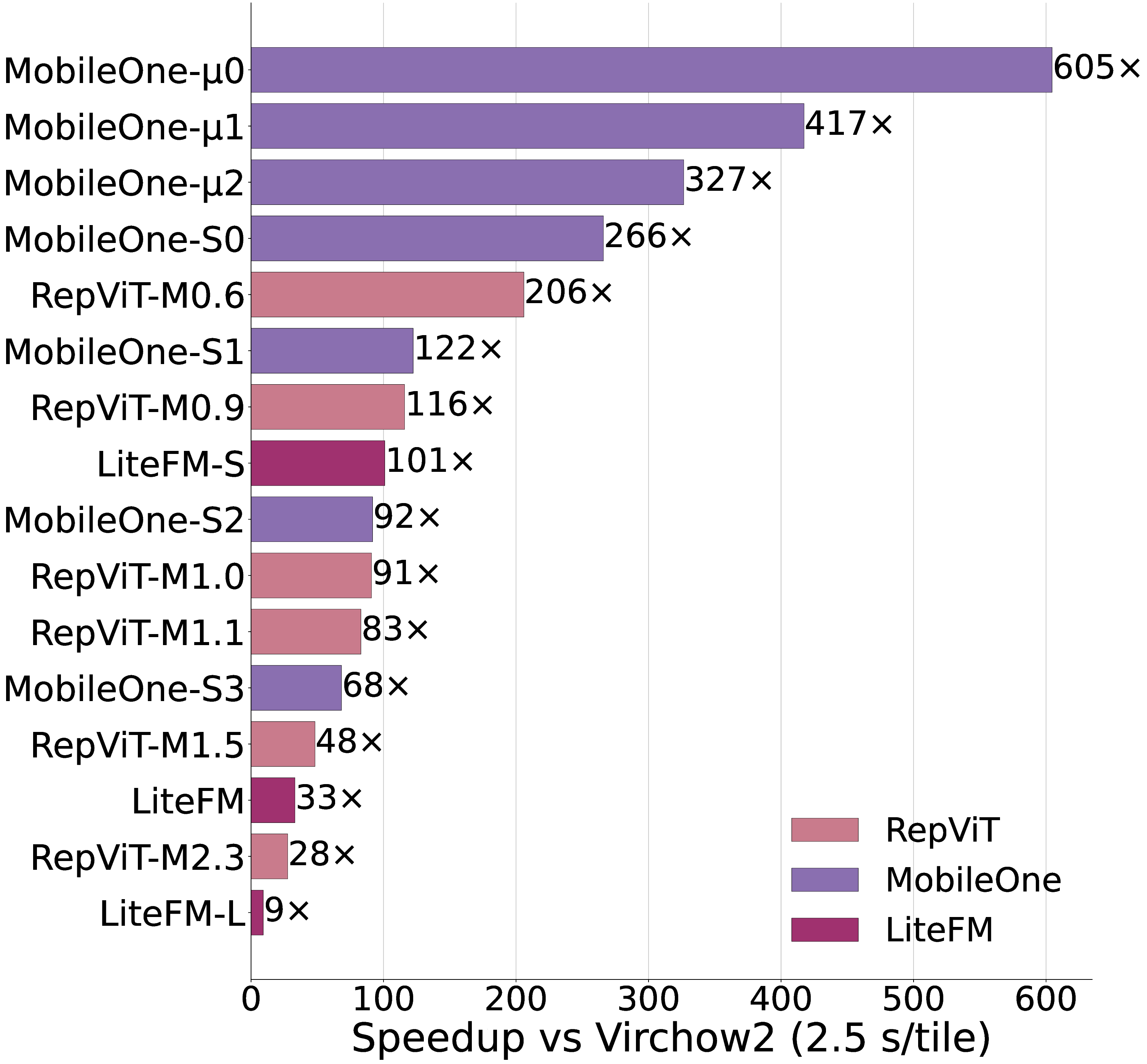} \\
    {\scriptsize\textbf{(a)} Performance-efficiency trade-off.} &
    {\scriptsize\textbf{(b)} Edge speedup on Raspberry Pi 5.}
  \end{tabular}
  \caption{\textbf{Efficiency and edge deployment.} \textbf{(a)} MuCoEdge models compared with LiteFM variants and teacher foundation models on CPTAC external validation. \textbf{(b)} Single-tile inference speedup over Virchow2 on Raspberry Pi 5.}
  \label{fig:efficiency_edge}
\end{figure}
The efficiency gains are substantial (Tab.~\ref{tab:feature_extractor_overview}, Fig.~\ref{fig:efficiency_edge}a). Speedups are reported against Virchow2, the most efficient high-performing teacher reference in our benchmark. MuCoEdge-R0.9 improves over LiteFM-L while using 18$\times$ fewer parameters, 21$\times$ fewer GFLOPs, and 12.9$\times$ faster Raspberry-Pi inference. MuCoEdge-R0.6 reaches 69.4\% AUROC with 2.2M parameters and runs 6.2$\times$ faster than LiteFM and 2.0$\times$ faster than LiteFM-S. The sub-million MobileOne variants provide the most aggressive latency reduction, achieving 327--605$\times$ single-tile speedups over Virchow2 while retaining 66.5--66.9\% external AUROC. Together, these results define a practical edge-deployment spectrum, from higher-accuracy RepViT models to sub-million-parameter MobileOne models for strict latency budgets.

\section{Conclusion}
MuCoDi transfers multiple PFM feature spaces into compact, edge-oriented pathology encoders. In broad internal TCGA and external CPTAC evaluation, MuCoEdge-R0.9 improves over LiteFM-L with 18$\times$ fewer parameters, MuCoEdge-R0.6 remains competitive with LiteFM at 2.2M parameters, and sub-million MobileOne variants retain 66.5--66.9\% external AUROC with 327--605$\times$ Raspberry-Pi speedups over Virchow2. From a clinical perspective, encoders that run directly on departmental workstations or microscope-attached edge devices remove the data-transfer and compute barriers that currently keep PFM-based decision support out of routine diagnostic workflows.

\begin{credits}
\subsubsection{\ackname}
The authors gratefully acknowledge computing-time support from the GWK through ZIH at TU Dresden and from the Gauss Centre for Supercomputing e.V. through NIC on the JUWELS and JUPITER Booster modules at Jülich Supercomputing Centre.
We also acknowledge the TCGA Research Network and the Clinical Proteomic Tumor Analysis Consortium (CPTAC), which generated the data on which the results shown in this study are based.
\subsubsection{\discintname} 
NGR received compensation for travel expenses from nanoString, a Bruker company. 
JNK holds shares in StratifAI, Synagen, Spira Labs, Tremont AI, and Saterra AI; is Co-PI on institutional research grants from GSK and AstraZeneca, and declares honoraria or consulting fees from AstraZeneca, Bayer, Bioptimus, Daiichi Sankyo, Eisai, Janssen, Merck, MSD, Novartis, BMS, Roche, and Pfizer.
\end{credits}

\bibliographystyle{splncs04}
\bibliography{mucodi_references}
\end{document}